\documentclass[runningheads]{llncs}
\usepackage[T1]{fontenc}
\usepackage{graphicx}
\begin{document}
\title{Russian-Language Multimodal Dataset for Automatic Summarization of Scientific Papers}
\titlerunning{Russian-Language Multimodal Dataset}
%
\author{Tsanda Alena\inst{1} \and Bruches Elena\inst{1,2}}
\authorrunning{Al. Tsanda et al.}
%

\institute{Novosibirsk State University, Russia \and A.P. Ershov Institute of Informatics Systems, Russia}

\maketitle              
\begin{abstract}
The paper discusses the creation of a multimodal dataset of Russian-language scientific papers and testing of existing language models for the task of automatic text summarization. A feature of the dataset is its multimodal data, which includes texts, tables and figures. The paper presents the results of experiments with two language models: Gigachat from SBER and YandexGPT from Yandex. The dataset consists of 420 papers and is publicly available on https://github.com/iis-research-team/summarization-dataset.

\end{abstract}
\section{Introduction}
It is essential for any researcher to read scientific literature in order to be aware of the latest trends and state-of-the-art solutions in their field of interest. However, with the wealth of textual content growing dramatically on the Internet, especially scientific papers, the reading process can be inconvenient. Therefore, researchers need representative abstracts to understand the topics of the papers and focus on the relevant information. Manual text summarization consumes a lot of time, effort and cost and even becomes impractical with the upward trend towards automation. The main goal of any automatic summarization system is to produce a summary that includes the main ideas of the input document in less space without repetitions~\cite{ref_article1}.

Existing solutions mainly work for the English language and for texts of the general domain. Besides, the methods of text summarization are not so developed for Russian, especially for domain-specific texts like scientific papers, hence the need to adapt such methods and develop new ones for this language. What is more, there are very few Russian-language datasets of scientific texts, especially for this task.

This paper is devoted to the development of a multimodal dataset of scientific papers written in Russian and testing existing language models on the collected data. The main specific feature of the dataset proposed in this work is its multimodality: it includes not only texts of papers and their abstracts, but also tables and figures with their descriptions. Singular focus on textual information fails to capture the full richness of multimodal data as most existing paper interpretation systems do~\cite{ref_article2}. Containing valuable information, tables and figures can noticeably improve the quality of abstracts. Moreover, the creation of such datasets is a long and painstaking process, but they are especially relevant due to the development of multimodal models. Multimodal models are the models that make predictions based on different modalities, i.e. different types of information~\cite{ref_article3}.

In this work, we have two main contributions:
\begin{enumerate}
    \item We present the carefully curated multimodal dataset for the scientific papers summarization for Russian, which covers 7 scientific domains and contains 420 scientific papers, including textual and visual information.
    \item We evaluate the most widely used language models for Russian, namely GigaChat and YandexGPT, on this dataset. The study showed that handling scientific texts is ambiguous for these models as a number of texts were rejected due to ethical reasons.
\end{enumerate}

\section{Related Works}

In this chapter, we review the literature about the task of summarization, existing datasets for this task and multimodal large language models.

\subsection{The Task of Summarization}

Large volumes of textual information on the Internet have complicated the task of searching and reading all the relevant documents. The field of automatic text summarization (ATS) has emerged as a solution for condensing extensive texts and creating accurate summaries~\cite{ref_article4}. ATS provides users with the opportunity to quickly grasp the main idea of a document without having to manually filter through information.

The main objective of any ATS system is to produce an effective summary, i.e. a summary that distills the most important information from a source (or sources) to produce an abridged version of the original information for a particular user(s) and task(s)~\cite{ref_article5}.

There is a wide variety of different classifications and applications of ATS systems. Such systems may be classified based on their approach, input size, summarization algorithm, summary content, summary type, summarization domain, etc. They may be applied in many different ways as well: news, sentiment, books, email summarization, legal or biomedical documents summarization and others~\cite{ref_article1}. Particularly, text summarization is extremely functional for scientific papers. Indeed, scholars often face the challenge of sifting through vast amounts of literature to keep up with developments in their field.

In terms of approaches, automatic text summarization is categorized into extractive, abstractive and hybrid~\cite{ref_article4}. The first approach is based on extracting the most high-scored sentences from the original document. Extractive approach is fast and simple, yet it produces summaries that are far from those written by humans. On the contrary, abstractive summarization is generated by either rephrasing or using the new words instead of simply extracting sentences from the initial document~\cite{ref_article6}, which makes it more like a human-written one.

Some challenges of ATS like finding the most informative segments of a text, summarizing long documents such as books, evaluating computer-generated summaries without the need for the human-produced summaries make this task one of the most arduous yet significant tasks of NLP.

\subsection{Datasets for Summarization}

Summarization aims to extract the most important information from the different types of texts. Thus, there are plenty of datasets serving the different tasks.The most popular summarization datasets are CNN/DailyMail~\cite{ref_article8} and XSum dataset [8], they are used for news summarization. Besides, there are task-specific datasets, such as for dialogue summarization~\cite{ref_article9,ref_article10}, chat summarization~\cite{ref_article11}, email summarization~\cite{ref_article12}.

The most challenging aspect is long text summarization. Intuitively, long document summarization is harder than short document one due to the significant difference in the number of lexical tokens and breadth of content between short and long documents. As the length increases, the content that would be considered important increases as well, resulting in a more challenging task for an automatic summarization model to capture all salient information in the limited output length~\cite{ref_article13}. Such texts may be presented in books, long documents and scientific papers. The most widely used datasets for scientific paper summarization are arXiv and PubMed~\cite{ref_article14}. However, these datasets are intended for English. As for the Russian language, there are very few datasets. For text summarization there is Gazeta~\cite{ref_article15}. Besides, there are Russian-language parts of the datasets MLSUM~\cite{ref_article16} and XL-Sum~\cite{ref_article17}.

Another important aspect of the current state-of-the-art is multimodal datasets, which contain not only text information, but also images, tables, etc. For example, in~\cite{ref_article18} SciMMIR benchmark was proposed to evaluate a model's abilities to use several types of information for information retrieval over the scientific texts. Constructing such datasets is a time-consuming process, but they may play an important role even in the model’s evaluation.

\subsection{Multimodal LLMs}

Large Language Models, or LLMs, are basically AI systems that use deep learning techniques and massively large datasets to comprehend and generate human language text. These models have transformed natural language processing (NLP), branching their influence into various domains~\cite{ref_article19}. Undoubtedly, LLMs with their unprecedented capabilities have redefined the way we understand language and generate text.

Along with a great deal of modern research, a wide variety of different LLMs have emerged. Large language models can be classified based on their size - from small ($\ge$ 1B parameters) to very large (> 100B parameters); on their type - foundation, instruction or chat model; on their origin and availability~\cite{ref_article20}. Besides, there are some very popular LLM families such as the GPT family developed by OpenAI~\cite{ref_article21}, the LLaMa family released by Meta~\cite{ref_article22}, the PaLM from Google~\cite{ref_article23} and others.

Operating mainly with the pure text data, traditional large language models perform well at tasks like text generation and encoding but have limitations in understanding other data types~\cite{ref_article24}. Meanwhile, large vision foundation models make rapid progress in perception. Consequently, more and more attention focuses on combination with text, modality alignment and task unity~\cite{ref_article25}. This leads to the new field of multimodal LLMs, or MLLMs, which handle various modalities of information. What is more, multimodality is a key component of achieving general artificial intelligence, because it plays a pivotal role in interacting with the real world.

The use of multimodal input vastly expands the capabilities of existing language models. However, it still remains an active area of research as the majority of LLMs are trained only on textual data.

\section{Dataset}

In this section, we provide a description of the dataset, the procedure of its creation and the statistics.

\subsection{Data Sources}

Our goal is to create a dataset for the scientific papers summarization for Russian. This dataset should cover different domains, including technical, humanitarian and natural sciences. Such diversity is important as each domain has its own scientific traditions, patterns and different set of metadata such as formulas, tables, figures etc. The created dataset demonstrates this diversity, which can be seen in Table~\ref{tab1} and Table~\ref{tab2}.

The following scientific domains were included in the dataset: Economics, History, Information Technologies (IT), Journalism, Law, Linguistics and Medicine.

The papers were collected from three scientific journals: Vestnik TSU\footnote{https://journals.tsu.ru/} (Economics, History, Journalism, Law, Linguistics, IT), Vestnik NSU\footnote{https://vestnik.nsu.ru/} (IT) and Medicinsky journal\footnote{https://sibmed.elpub.ru/jour/index} (Medicine).

It should be noticed that even within the same journal and domain the structure of the papers can vary significantly. For example, there is no unified structure for the Journalism papers: some texts are divided by sections while others are continuous texts without any explicit structure.

We consider this feature as a challenge for the modern language models – in practice, we expect that they will extract the most important information without memorizing or relying on the section names, structure and other features that can be used as a hint for the models.

\subsection{Dataset Creation}

The dataset consists of the papers from 7 different scientific domains, each domain has 60 texts.

All texts were collected from the newest publications to the earlier ones. We chose only papers which contain the most part of the text in Russian. Some papers contain quite a lot of text inclusions in other languages such as Chinese, German, etc.

Each text has the following metadata:
\begin{enumerate}
    \item Name of the paper;
    \item Abstract;
    \item Text of the paper;
    \item Figures and their names;
    \item Tables and their names.
\end{enumerate}

It is worth noting that we clean text from such metadata as author names, keywords, footnotes, references and appendix. Additionally, we remove all links to the references from the text.

All figures and tables contained in the papers were screenshotted and saved as .png files. The descriptions for the visual data were extracted from their headings in the paper. In case there was no heading for the figure or table, this information was extracted from the context. For example, from the context “\textit{We show the details of TIVE in Fig. 2}” one can extract the possible description for the figure  “\textit{The details of TIVE}”.

Some tables may be torn into two or more pages. In this case, we save the tables in several .png files, depending on the number of pages containing this table.

In terms of the dataset storage, each paper occupies one folder, which contains the following files: name.txt, abstract.txt, text.txt, image\_number, table\_number, figures.json, tables.json.

\subsection{Dataset Statistics}

As it was mentioned before, the current version of the dataset consists of 420 texts from 7 scientific domains.

Tables~\ref{tab1} and \ref{tab2} show the general statistics such as text length, number of tokens, figures and tables for the abstracts and texts correspondingly.

One may notice that in humanitarian domains texts tend to be longer, but they rarely include figures and tables. Technical and natural sciences contain the most part of visual information and shorter texts. This insight shows that quite a lot of useful information is encoded in other modalities than textual one and may play a crucial role for such semantic tasks as summarization, information extraction, etc.

\begin{table}
\caption{Statistics for the abstracts.}\label{tab1}
\centering
\begin{tabular}{|l|l|l|}
\hline
Domain &  Length in chars & Length in tokens\\
\hline
Economics &  64 271 & 7 004\\
History &  34 211 & 3 787\\
IT & 36 277 & 3 822\\
Journalism & 31 981 & 3 664\\
Law & 33 288 & 3 423\\
Linguistics & 35 190 & 3 806\\
Medicine & 97 061 & 11 296\\
Total & 332 279 & 36 802\\
\hline
\end{tabular}
\end{table}

\begin{table}
\caption{Statistics for the texts.}\label{tab2}
\centering
\begin{tabular}{|l|l|l|l|l|}
\hline
Domain &  Length in chars & Length in tokens & Figures & Tables\\
\hline
Economics &  1 316 995 & 151 284 & 32 & 25\\
History &  1 540 251 & 184 407 & 2 & 17\\
IT & 1 002 115 & 114 721 & 238 & 27\\
Journalism & 1 377 087 & 174 064 & 45 & 12\\
Law & 1 243 153 & 143 675 & 0 & 2\\
Linguistics & 1 557 481 & 190 478 & 1 & 1\\
Medicine & 963 178 & 107 449 & 19 & 45\\
Total & 9 000 260 & 1 066 078 & 337 & 129\\
\hline
\end{tabular}
\end{table}

\section{Benchmarks}

In this section, we describe the used methods and the obtained results.

\subsection{Models}

As a baseline, we decided to check whether the final section of the papers in the dataset correlates with the abstracts. Due to the fact that the papers do not always have a strict structure, not every paper has the section "Conclusion". In total, 183 pairs of "abstract – conclusion" were formed and taken to evaluate the quality. The results are presented in Tables~\ref{tab3} and \ref{tab4} in section 4.2.

Besides, we conducted some experiments with large language models. The first model whose performance was tested is Gigachat\footnote{https://developers.sber.ru/portal/products/gigachat} from SBER. This model is based on a neural network ensemble which includes, in particular, models for text generation. The significant advantage of this language model is that it was trained for the Russian language. However, Gigachat strives to avoid controversial ethical issues, and for this reason only 37\% of texts from the dataset were processed without being censored. The model is particularly reluctant to work with legal and medical texts, as well as topics related to journalism. Nevertheless, 157 abstracts to dataset papers were generated with help of the langchain framework\footnote{https://github.com/langchain-ai/langchain} designed to create applications using large language models. We evaluated the quality of generated abstracts taking the abstracts from the dataset as the reference summaries just as we did in the baseline. The prompt used for addressing the model: “\textit{Below is a scientific paper. Highlight the main facts and write a summary of this paper}”.

The second model that has been utilized is the generative language model from Yandex called YandexGPT\footnote{https://cloud.yandex.ru/ru/services/yandexgpt}. Like the previous one, this model was trained on Russian-language data. With help of the Yandex GPT Lite model, we generated 295 abstracts for the dataset papers using the exact same prompt as earlier. Along with the previous one, this model did not manage to process all 420 papers due to censorship and length of some texts. All Yandex GPT API generation models currently have a limit of 8 000 tokens for the input and output sequences together.

\subsection{Results}

To evaluate the quality of the generated abstracts, it is required to compare the candidate summaries with the reference ones, i.e. the ones from the dataset, in order to determine how equivalent they are. For this purpose, the following text generation metrics were selected: BERTScore, BLEURT, ROUGE-1, ROUGE-2, ROUGE-L, BLEU. It is worth mentioning that expert evaluation of the collected dataset was not carried out.

The first metric, BERTScore, is based on pre–trained contextual embeddings BERT and calculates the similarity of sentences as the sum of cosine similarities between the embeddings of the tokens they consist of~\cite{ref_article26}. This metric is especially effective for abstractive summarization since it takes into account such points as paraphrasing and changing the order of words.

The second metric, BLEURT, similarly to the first one uses BERT embeddings and takes into account the semantics of the text. A key component of this metric is a special pre-training scheme on a large volume of synthetic data~\cite{ref_article27}.

The ROUGE and BLEU metrics are considered to be standard for text generation and are based on calculating overlapping n-grams. They are more suitable for extractive summarization, since such summarization is formed from the most important sentences of the source document. ROUGE-N is a metric related to recall, while BLEU is related to accuracy~\cite{ref_article28}. Apart from this difference, the BLEU metric uses a brevity penalty so that the candidate text matches the reference one not only in the choice and order of words, but also in length~\cite{ref_article29}. Both metrics calculate the percentage of n-grams in the candidate summary that match the n-grams in the reference summary, i.e. abstracts from our dataset.

Except BLEURT, the values of the metrics range from 0 to 1, where 1 means the best match. The BLEURT score may go beyond the specified range and be negative, but the perfect value is also considered to be 1.

During experiments with the large language models, the first step was to calculate the values of these metrics on those abstracts that the models managed to generate (Table~\ref{tab3}).

Unfortunately, it is quite problematic to compare the values of the metrics with one another in this case, since the number of generated texts, their lengths and domains differ significantly. To take into account the number of generated summaries and plausibly compare the work of the models, the same metrics were calculated for all 420 abstracts, including empty lines if the model did not manage to generate a summary (Table~\ref{tab4}). Although there was a substantial decline in values of the metrics in this case, it became possible to compare the models within this particular task. Additionally, we evaluated the models performance across different domains using the metrics: BERTScore, BLEURT, ROUGE-1, ROUGE-2, ROUGE-L (Table~\ref{tab5}).

The first LLM, Gigachat, performed the worst according to the metrics. The reason for this might be that the model managed to generate the fewest number of summaries due to censorship,  almost half as many as the YandexGPT model. For instance, the model generated only 9 abstracts for Journalism out of 60 possible due to these restrictions, yet it processed almost all IT papers. Unlike the model of Yandex, its considerable advantage is that it works with longer papers.

On the contrary, YandexGPT has the best scores for almost all the metrics. While this model does have some thematic limitations, it has processed 40 papers for Journalism. However, due to the length of the texts, it generated the fewest number of summaries for scientific fields like Linguistics and History.

In addition, the abstracts generated by the models differ in length. The average summary from Gigachat contains about 93 words or 5 sentences, while the average summary generated by YandexGPT has approximately 218 words or 15 sentences. This makes the latter significantly longer and may affect the metrics.

Some examples of the generated texts are provided here: https://github.com/iis-research-team/summarization-dataset/blob/main/generated\_abstracts.md.

\begin{table}
\caption{Metrics on the generated abstracts.}\label{tab3}
\centering
\begin{tabular}{|l|l|l|l|l|l|l|}
\hline
Model &  BERTScore & BLEURT & ROUGE-1 & ROUGE-2 & ROUGE-L & BLEU\\
\hline
Baseine &  0.708 & 0.187 & \textbf{0.173} & \textbf{0.099} & \textbf{0.167} & 1.4e-155\\
Gigachat &  \textbf{0.719} & \textbf{0.189} & 0.148 & 0.071 & 0.142 & 1.3e-155\\
YandexGPT & 0.692 & 0.204 & 0.118 & 0.053 & 0.113 & 5.1e-80\\
\hline
\end{tabular}
\end{table}

\begin{table}
\caption{Metrics on all the abstracts.}\label{tab4}
\centering
\begin{tabular}{|l|l|l|l|l|l|l|}
\hline
Model &  BERTScore & BLEURT & ROUGE-1 & ROUGE-2 & ROUGE-L & BLEU\\
\hline
Baseine &  0.310 & -0.295 & 0.075 & \textbf{0.044} & 0.073 & 1.3e-155\\
Gigachat &  0.270 & -0.349 & 0.055 & 0.027 & 0.053 & 0.9e-155\\
YandexGPT & \textbf{0.486} & \textbf{-0.057} & \textbf{0.083} & 0.037 & \textbf{0.080} & 5.1e-80\\
\hline
\end{tabular}
\end{table}

\begin{table}
\caption{Metrics by domains.}\label{tab5}
\centering
\begin{tabular}{|l|l|l|l|l|l|l|}
\hline
Model & Texts & BERTScore & BLEURT & ROUGE-1 & ROUGE-2 & ROUGE-L \\
\hline
Baseine\textsubscript{\textit{Medicine}} & 52 & \textbf{0.725} & 0.143 & 0.263 & \textbf{0.191} & \textbf{0.251}\\
Gigachat\textsubscript{\textit{Medicine}} & 39 & 0.718 & 0.158 & 0.203 & 0.116 & 0.194\\
YandexGPT\textsubscript{\textit{Medicine}} & \textbf{53} & 0.718 & \textbf{0.172} & \textbf{0.268} & 0.128 & 0.248\\
\hline
Baseline\textsubscript{\textit{Economics}} & 26 & 0.699 & 0.210 & \textbf{0.030} & \textbf{0.005} & \textbf{0.026}\\
Gigachat\textsubscript{\textit{Economics}} & 13 & \textbf{0.740} & \textbf{0.251} & 0.0 & 0.0 & 0.0\\
YandexGPT\textsubscript{\textit{Economics}} & \textbf{45} & 0.686 & 0.226 & 0.017 & 0.0 & 0.017\\
\hline
Baseline\textsubscript{\textit{Linguistics}} & 11 & 0.687 & \textbf{0.207} & \textbf{0.123} & \textbf{0.109} & \textbf{0.123}\\
Gigachat\textsubscript{\textit{Linguistics}} & 22 & \textbf{0.718} & 0.182 & 0.034 & 0.0 & 0.034\\
YandexGPT\textsubscript{\textit{Linguistics}} & \textbf{37} & 0.685 & 0.199 & 0.067 & 0.022 & 0.064\\
\hline
Baseline\textsubscript{\textit{Law}} & 9 & 0.678 & 0.220 & 0.0 & 0.0 & 0.0\\
Gigachat\textsubscript{\textit{Law}} & 14 & \textbf{0.718} & \textbf{0.253} & 0.029 & 0.0 & 0.029\\
YandexGPT\textsubscript{\textit{Law}} & \textbf{39} & 0.684 & 0.226 & \textbf{0.033} & \textbf{0.009} & \textbf{0.033}\\
\hline
Baseline\textsubscript{\textit{IT}} & 55 & 0.720 & 0.205 & \textbf{0.210} & \textbf{0.105} & \textbf{0.204}\\
Gigachat\textsubscript{\textit{IT}} & 50 & \textbf{0.734} & \textbf{0.209} & 0.178 & 0.098 & 0.169\\
YandexGPT\textsubscript{\textit{IT}} & \textbf{57} & 0.691 & 0.203 & 0.114 & 0.047 & 0.111\\
\hline
Baseline\textsubscript{\textit{Journalism}} & 22 & 0.687 & \textbf{0.212} & 0.149 & 0.029 & 0.149\\
Gigachat\textsubscript{\textit{Journalism}} & 9 & \textbf{0.702} & 0.084 & \textbf{0.262} & \textbf{0.089} & \textbf{0.262}\\
YandexGPT\textsubscript{\textit{Journalism}} & \textbf{40} & 0.683 & 0.205 & 0.134 & 0.044 & 0.134\\
\hline
Baseline\textsubscript{\textit{Chemistry}} & \textbf{56} & \textbf{0.755} & \textbf{0.190} & \textbf{0.411} & \textbf{0.244} & \textbf{0.381}\\
Gigachat\textsubscript{\textit{Chemistry}} & 41 & 0.728 & 0.172 & 0.252 & 0.121 & 0.242\\
YandexGPT\textsubscript{\textit{Chemistry}} & \textbf{56} & 0.713 & 0.142 & 0.243 & 0.096 & 0.219\\
\hline
Baseline\textsubscript{\textit{History}} & 9 & 0.674 & 0.158 & 0.108 & 0.069 & 0.108\\
Gigachat\textsubscript{\textit{History}} & 10 & \textbf{0.704} & 0.206 & \textbf{0.281} & 0.100 & \textbf{0.281}\\
YandexGPT\textsubscript{\textit{History}} & \textbf{24} & 0.689 & \textbf{0.210} & 0.179 & \textbf{0.136} & 0.179\\
\hline
\end{tabular}
\end{table}

\section{Limitations}

Current version of the dataset covers only 7 scientific domains, but it should be increased to include more diverse areas.

The most challenging part is to create a subset with technical papers such as Maths or Physics as they contain plenty of formulas that are important data. It is still an open question how one should store such information: as a raw text or as a LateX text. We are going to solve this problem in our future work.

Moreover, current evaluation does not support all modalities (both text and images) as input for the LLMs as only few models support such API. This is also a very important direction of our ongoing research. Nevertheless, this dataset may already be used to evaluate the systems that support multimodal inputs.

Another nuance is table representations as images. However, it seems that one can make use of processing tabular data (for example, in csv format) as such modality is included in some modern models.

\section{Conclusion}

In conclusion, we proposed a russian-language multimodal dataset for the task of automatic text summarization and conducted some experiments that include testing large language models. The dataset is publicly available on GitHub, and we plan to expand it with other scientific fields.

According to the traditional text generation metrics, the final sections in the scientific papers of the corpus turned out to be syntactically closer to the abstracts in the corpus than those generated by the models. On the contrary, in terms of semantics large language models performed better according to the neural network metrics, which take into account the content of the text. Despite the fact that LLMs today are capable of generating high-quality text in natural language, they have some limitations concerning the length and the content of texts. 

Currently, we plan to test the performance of other language models on the resulting dataset and conduct experiments with different extractive and abstractive approaches to text summarization. Through combining methods and analyzing the collected figures and tables, we aim to enhance the quality of the generated abstracts.

%
%
%
%

\end{document}